\documentclass{article}

\usepackage{xcolor}         % colors; load before corl_2026 hyperlink colors
\usepackage[preprint]{corl_2026} % Uncomment for pre-prints (e.g., arxiv); This is like ``final'', but will remove the CORL footnote.

\title{Is Your Trajectory Displacement Safe in Long-tail?}
\usepackage[utf8]{inputenc} % allow utf-8 input
\usepackage[T1]{fontenc}    % use 8-bit T1 fonts
\usepackage{hyperref}       % hyperlinks
\usepackage{url}            % simple URL typesetting
\usepackage{booktabs}       % professional-quality tables
\usepackage{tabularx}       % flexible-width tables
\usepackage{amsfonts}       % blackboard math symbols
\usepackage{amsmath}        % advanced math formatting
\usepackage{nicefrac}       % compact symbols for 1/2, etc.
\usepackage{microtype}      % microtypography
\usepackage{graphicx}       % figures
\usepackage{float}          % fixed figure placement with [H]
\usepackage{enumitem}       % list formatting

\usepackage{pifont}

\newcommand{\NATR}{No Additional Threat Rate (NATR)}
\newcommand{\NATRShort}{NATR}

\usepackage{longtable}
\usepackage{array}
\usepackage{tikz}
\usetikzlibrary{arrows.meta,positioning,shapes.geometric}

\author{
    Qiao Sun$^{1}$ \qquad
    Weicheng Zheng$^{1,3}$ \qquad
    Yixin Huang$^{1,3}$ \qquad
    Hang Zhao$^{1,2}$\thanks{Corresponding author: hangzhao@mail.tsinghua.edu.cn} \\
    \\
    $^1$Shanghai Qi Zhi Institute \qquad
    $^2$Tsinghua University \qquad
    $^3$Tongji University
}

\begin{document}
\maketitle

% % Two or three meaningful keywords should be added here
% \keywords{CoRL, Robots, Learning}

\begin{abstract}
  Long-tail scenarios remain a major bottleneck for autonomous driving evaluation, even as datasets grow by orders of magnitude. Existing evaluation pipelines are rarely \textbf{human-aligned, safety-aware, verifiable, and explainable} at the same time: closed-loop metrics often saturate among strong planners, while unstructured human ratings can be noisy without a carefully designed protocol. We formulate planning evaluation as additional-threat detection: given a planner trajectory and an expert reference, does the planner's displacement introduce new unsafe driving behavior? We propose \textbf{FluidTest}, an evaluation pipeline with three components: a pairwise WebUI protocol for reliable human annotation; a taxonomy of 32 semantic threats with evidence-grounded decision graphs; and a three-agent verification system with reflection for precision and auditability. 
  % Experiments show that FluidTest produces consistent labels across annotators and planners, and identifies additional threats in \textbf{51\%} of RAP trajectories and \textbf{65\%} of Poutine trajectories. 
  Experiments on the WOD-E2E show that FluidTest produces consistent labels among trained annotators, and identifies additional threats in \textbf{65\%} of Poutine, and \textbf{51\%} of RAP trajectories.
  These results show that state-of-the-art planners can still exhibit substantial safety-relevant failures despite high Rater Feedback Scores (RFS) and low average displacement error (ADE). 
%   Additional details are available on our reviewer website: \url{https://safety-arena-web.vercel.app}.

\end{abstract}


\section{Introduction}

Scaling has improved autonomous driving planning across much of the data distribution, including some tail scenarios~\citep{naumann2025data,sun2025generalizing}. As performance on average cases improves, residual risk increasingly concentrates in the deepest tail~\citep{hallgarten2024interplan,okelly2018rareevent}. A common evaluation approach is to run closed-loop simulations initialized from real-world scenarios~\citep{caesar2021nuplan,dauner2024navsim,Cao2025CORL}, using rule-based checks and aggregate metrics. This pipeline is efficient but limited.

\begin{figure*}[t]
  \centering
  \includegraphics[width=1\linewidth]{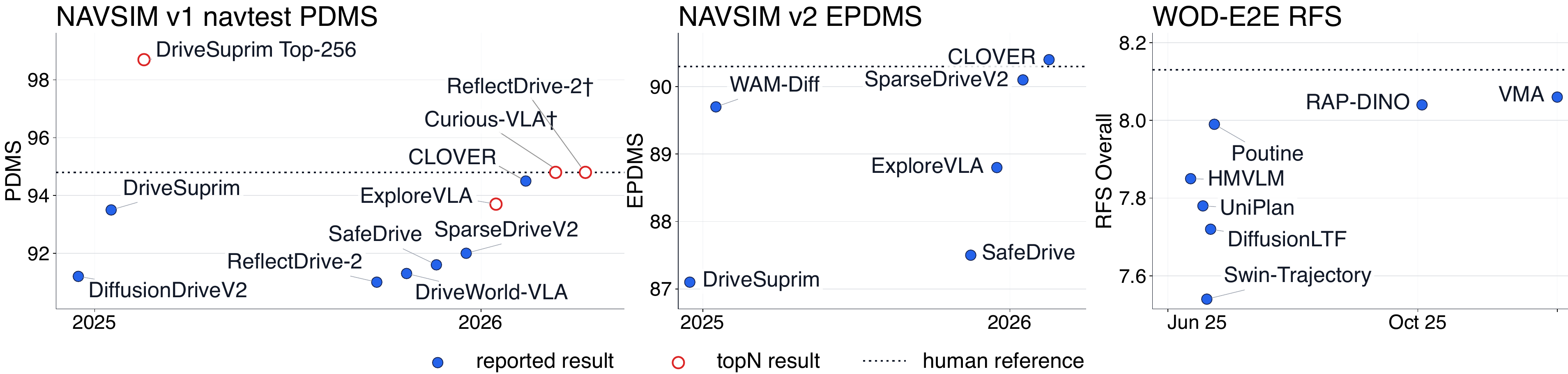}
  % \vspace{-9pt}
  \caption{Existing benchmarks are saturated, making it difficult to distinguish among planners and potentially obscuring safety-relevant failures in long-tail driving scenarios. The x-axis indicates the first release date of each method's report.}
  \label{fig:saturated-metrics}
  % \vspace{8pt}
\end{figure*}

First, closed-loop metrics are not always aligned with human driving preferences~\citep{dauner2023parting}, creating interaction-level sim-to-real gaps: a planner may accelerate instead of yielding and still avoid a simulated collision, even though the behavior would be unsafe or undesirable in real traffic. Second, closed-loop benchmarks require accurate 3D reconstruction of maps, traffic controls, and agent corridors to measure events such as red-light violations and collisions, which limits their applicability in long-tail scenes with complex topology, occlusions, or poor lighting. Third, existing metrics often reduce safe driving to a narrow set of event checks, making them too coarse for real-world driving semantics~\citep{jia2024bench2drive}. They can also be vulnerable to reward hacking and score saturation, as shown in Fig.~\ref{fig:saturated-metrics}.

These limitations motivate benchmarks such as WOD-E2E, the Waymo Open Dataset for End-to-End Driving in Challenging Long-tail Scenarios~\citep{xu2025wode2e}. WOD-E2E covers diverse challenging scenarios, and its \textbf{Rater Feedback Score (RFS)} captures human preferences. However, its current testing pipeline does not use the expert trajectory as a reference, even though expert behavior provides crucial scene context~\citep{rowe2025poutine}. Our experiments identify three limitations: scored samples are sparse relative to the output space of modern planners; scalar scores can be unreliable across planners and scenes; and the scores lack explanations for failure diagnosis.

% include Poutine, RAP, and an internal high-performing planner, denoted VMA
% \footnote{
%   VMA achieves state-of-the-art performance on the official WOD-E2E test-set benchmarks for both RFS (8.06) and 5sADE (2.61).
%   VMA is included only as an additional stress test. The main conclusions also hold for the public planner families Poutine and RAP.
%   % The implementation details of VMA are outside the scope of this paper; we refer readers to the forthcoming VMA report for technical details.
% }, and find that RFS has three limitations: scored samples are sparse relative to the output space of modern planners; scalar scores can be unreliable across planners and scenes; and the score lacks explanations for failure diagnosis.

\begin{figure*}[t]
  \centering
  \includegraphics[width=\textwidth]{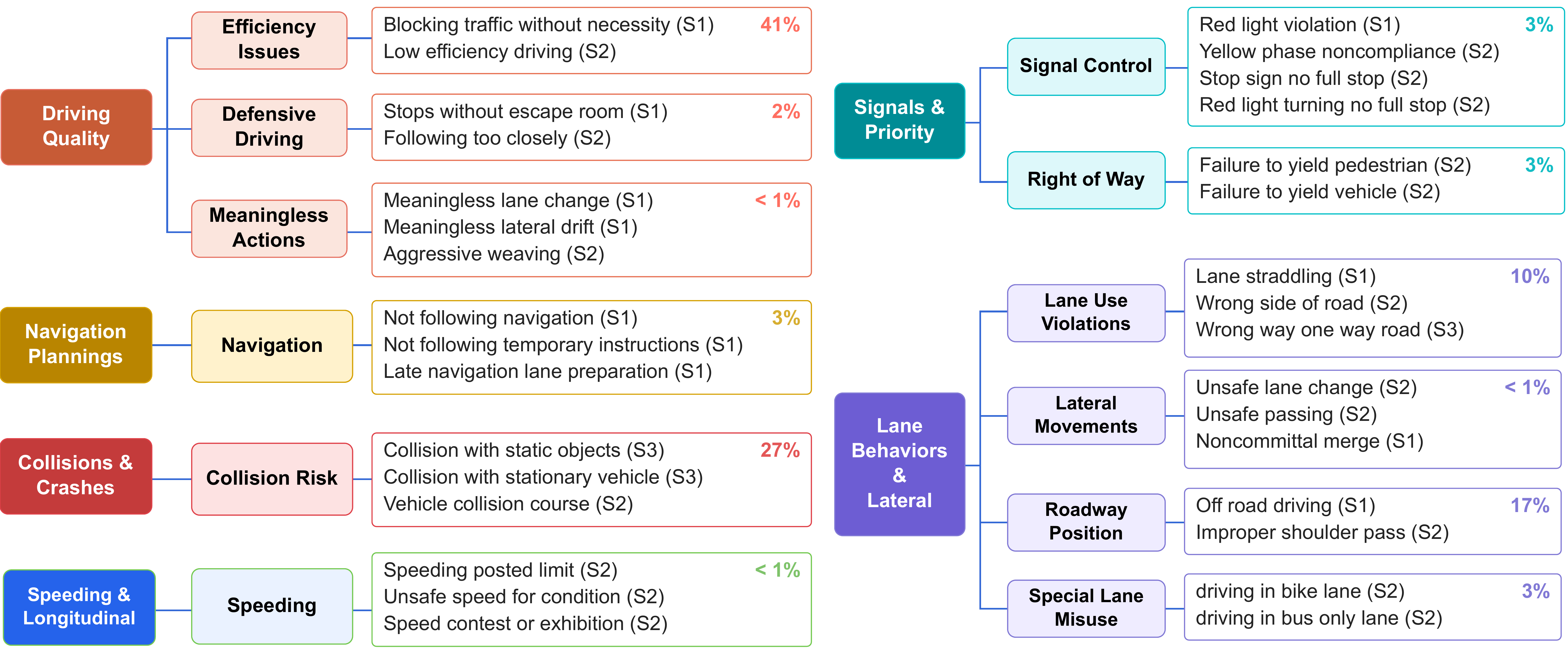}
  % \vspace{-8pt}
  \caption{Threat taxonomy used in FluidTest. The taxonomy is substantially more comprehensive than NAVSIM v1 (5 metrics) and NAVSIM v2 (9 metrics). The percentage in each box indicates the violation rate for the Poutine planner on the WOD-E2E val151 subset.}
  \label{fig:threat-taxonomy}
\end{figure*}

We argue that manually designed metrics such as PDMS introduce inductive biases that can weaken reliability on hard cases (Sec.~\ref{sec:hard-case-universality}). Since pairwise comparisons are generally easier for annotators than scalar ratings~\citep{kiritchenko2017best}, we ask annotators to directly answer: \emph{does the planner's displacement from the expert trajectory introduce an additional unsafe driving behavior?} Each displacement is classified as \texttt{No Threat}, \texttt{Has Threat}, or \texttt{Not Sure}. We further organize unsafe driving behavior into \textbf{32} semantic threat categories, parsed from the California DMV's \emph{Driver Handbook}~\citep{cadmv2026handbook}, and define \textbf{\NATR{}} as the main metric. We then build \textbf{FluidTest}, a three-agent verification system with a reflection, revision loop to improve reliability, precision, and explainability than other existing benchmarks.

For evaluation, we filter the WOD-E2E validation set using three criteria to reduce noise, yielding 151 scenarios\footnote{
  Following the WOD-E2E setting, each scene corresponds to the critical frame extracted from a long driving log. This subset is already large compared with the 136 non-long-tail logs in \texttt{navtest}~\citep{dauner2024navsim}.
}: no obvious future uncertainty, no trivial constant-velocity expert plan, and no severe camera-parameter errors for trajectory projection. We benchmark two state-of-the-art planners with diverse implementations: Poutine\footnote{
  We use a self-reimplemented version because no official code or checkpoint has been released. Our implementation achieves similar RFS (7.89) and 5sADE (2.77), matching Poutine-Base on the test set.
}~\citep{rowe2025poutine}, a VLM-based planner, and RAP~\citep{feng2025rap}, a classical end-to-end planner, as well as an internal high-performing planner, denoted VMA\footnote{
  VMA achieves SOTA performance on the official WOD-E2E benchmarks for both RFS (8.06) and 5sADE (2.61).
  VMA is included only as an additional stress test. 
  % The main conclusions also hold for the public planner families Poutine and RAP.
  % The implementation details of VMA are outside the scope of this paper; we refer readers to the forthcoming VMA report for technical details.
}. Our WebUI and labeling protocol produce consistent labels across planners, with an overlap rate of 79.31\% and a Fleiss' $\kappa$ value of 0.7145 among researcher annotators. The final evaluation results show that both average displacement error (ADE) and RFS are misaligned with human preferences as measured by \NATRShort{}.

Our contributions are threefold:
% \begin{enumerate}[leftmargin=12pt]
\begin{enumerate}
\item We show that saturated PDMS and RFS fail to capture complex planning threats, and we propose \textbf{\NATR{}} for human-aligned safety evaluation.
\item We introduce \textbf{FluidTest}, a safety-aware, explainable, and comprehensive protocol for benchmarking planning trajectory quality in long-tail scenarios.
\item We release a WebUI and leaderboard for benchmarking new planners, along with labeling code, training code, checkpoints for the threat-detection Prosecutor model, and core multi-agent testing components, including definitions and prompts.
\end{enumerate}

\begin{figure}[t]
  \centering
  \includegraphics[width=\linewidth]{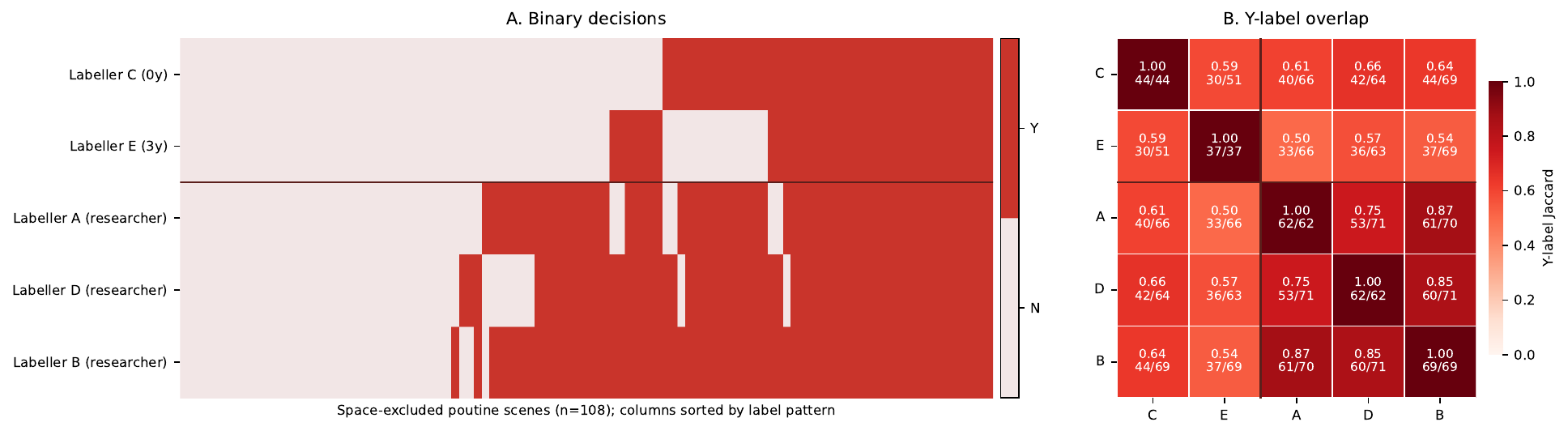}
  % \vspace{-10pt}
  \caption{Heatmaps of human labelers' results on the val151 set using Poutine's planning outputs. The researcher-group results appear at the bottom of the left panel and the lower right of the right panel, showing high overlap and strong labeling consistency. Scenarios labeled "Unsure" are excluded.}
  \label{fig:human_consistency_heatmaps}
\end{figure}

\section{Evaluations for Long-tail Planning}

Autonomous driving planners should perform at least as safely and reliably as human drivers. Yet current planners can still underperform humans in long-tail scenarios, and the observed worst-case distribution can depend strongly on the objectives used during optimization and evaluation. We argue that existing evaluation pipelines are both misaligned with human driving preferences and insufficiently discriminative for separating strong planners from weak ones.

\paragraph{Simulation benchmarks are saturated.}
Closed-loop simulation benchmarks aim to align planner evaluation with driving safety, but several limitations prevent them from fully reflecting real-world complexity. Many rely on hand-crafted or limited scenario sets~\citep{dosovitskiy2017carla,jia2024bench2drive}; use simplified behavior models~\citep{caesar2021nuplan} or non-reactive log replay~\citep{dauner2024navsim}; and reduce safe driving to a small set of predefined events. These choices can create unrealistic interaction patterns, including an implicit ``others always yield'' bias. As a result, planner scores can quickly saturate near or even above reported human-level performance, leaving dangerous long-tail planning failures undetected.

\paragraph{Human rater feedback is useful but not sufficient.}
Improvements on average-case scenarios do not necessarily reduce rare but safety-critical failures~\citep{hallgarten2024interplan,okelly2018rareevent}. WOD-E2E~\citep{xu2025wode2e} addresses this issue by collecting challenging long-tail driving segments and using human rater feedback to evaluate end-to-end planning outputs. However, the evaluation protocol becomes the next bottleneck: although RFS is annotated by human raters, our experiments show that it does not consistently reflect human safety preferences and provides limited diagnostic information.

\paragraph{VLMs as verifiers and simulators.}
A practical long-tail planning protocol requires both simulation and verification. Accurate simulation in complex interactive scenarios can be as difficult as the planning problem itself~\citep{sun2022intersim}. 
% Recent progress suggests that VLMs and LLMs can serve as useful world models
Recent progress suggests that vision-language models (VLMs) and large language models (LLMs) can serve as useful world models~\citep{li2025drivevlaw0, liang2025eurekaverse}. 
Additonally, recent LLM-symbolic systems show that LLMs are more reliable when they translate unstructured inputs into structured programs, logical forms, or planning representations, while symbolic modules perform execution, inference, or validation~\citep{pan2023logic,gao2023pal,liu2023llmp}. This motivates our design: FluidTest uses VLM/LLM agents for semantic scene understanding and threat proposal, but constrains final decisions through evidence-grounded decision graphs. This hybrid design is especially suitable for driving, where neural perception must be combined with interpretable rule and safety reasoning~\citep{sun2021neuro}.

% On NAVSIM v1, recent planners reach scores of 90.8~\citep{li2025recogdrive}, 93.0~\citep{li2025drivevlaw0}, and 98.7~\citep{yao2026drivesuprim}, compared with a human driving score of 94.8~\citep{dauner2024navsim}. On NAVSIM v2, planners reach 87.1~\citep{yao2026drivesuprim}, 88.8~\citep{sheng2026explorevla} and 90.4~\citep{ang2026clover}. On Bench2Drive, planners reach 95~\citep{nguyen2025lead}, compared with 91.85 for the privileged RL planner Think2Drive~\citep{li2024think2drive} and 97.02 for a carefully curated privileged driver~\citep{jia2024bench2drive}. These saturated scores make marginal gains on existing benchmarks less informative and only weakly connected to real-world performance differences at production scale.

% In this paper, we find that large VLMs can understand and identify threats but cannot reliably output without explicit decision graphs and boundary definitions. As a result, VLMs embedded in an agentic system can provide reasonable simulation-like explanations for analyzing threats in long-tail driving scenarios.

\begin{figure*}[t]
  \centering
  % \vspace{-10pt}
  \includegraphics[width=\textwidth]{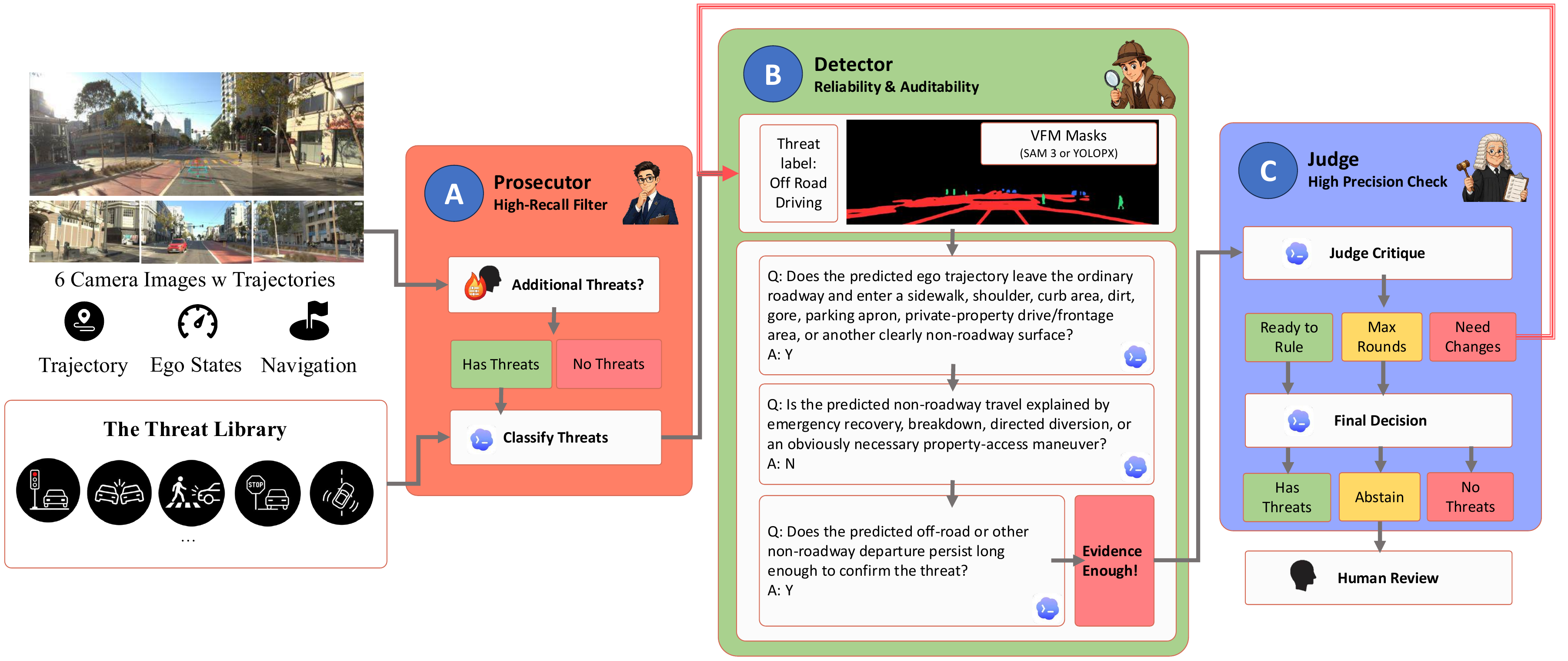}
  % \vspace{-10pt}
  % \caption{Overview of the FluidTest pipeline. Human labellers, or a VLM, label additional threats. Codex powers a multi-agent critique-and-repair loop that resolves ambiguous cases before producing the final confirmed threat set with decision graphs and detailed explanations. Finally, for a small group of abstained cases, human reviewers double-check the results and make the final decisions.}
  \caption{Overview of the FluidTest pipeline. Human annotators, or a VLM, label additional threats. Codex powers a multi-agent critique-and-repair loop that resolves ambiguous cases before producing a confirmed threat set with decision graphs and explanations. Human reviewers make final decisions for abstained cases.}
  \label{fig:main-pipeline}
\end{figure*}

\section{FluidTest}

\subsection{Reasoning and Metrics}
\label{sec:threat-definitions}

Driving in complex long-tail scenarios requires a planner to avoid risky maneuvers, even when they do not cause an immediate collision within the planning horizon. This makes explicit simulation both expensive and insufficient, so FluidTest directly evaluates whether a planner trajectory introduces additional threats relative to the expert trajectory.

We define \NATRShort{} using a taxonomy of 32 semantic threats, as shown in Fig.~\ref{fig:threat-taxonomy}, Appendix~\ref{sec:appendix-threats-def}, and Appendix~\ref{sec:appendix-threats-def}. These threats cover deterministic failures, interaction risks, roadway and lane-use errors, route-compliance failures, and low-quality driving behaviors. Let $N=|\mathcal{S}|$ be the number of evaluated scenes. For each scene $s\in\mathcal{S}$, FluidTest outputs confirmed threat labels $Y_s\subseteq\mathcal{L}$, where $\mathcal{L}$ is the full threat set. For any threat subset $A\subseteq\mathcal{L}$, we define
\[
z_s(A)=\mathbf{1}\!\left[Y_s\cap A\neq\emptyset\right],
\qquad
r(A)=\frac{1}{N}\sum_{s\in\mathcal{S}} z_s(A),
\qquad
% \mathrm{Score}(A)=1-r(A).
\mathrm{NATR}(A)=1-r(A).
\]
Here, $r(A)$ is the violation rate for threat set $A$, and $\mathrm{NATR}(A)$ is the corresponding no-violation score. The overall \NATRShort{} is obtained by setting $A=\mathcal{L}$.

\begin{figure}[t]
  \centering
  \includegraphics[width=\linewidth]{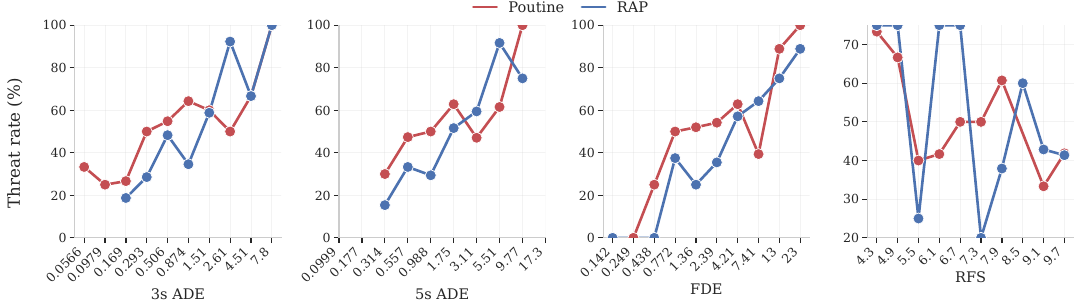}
  % \vspace{-15pt}
  \caption{Threat rate versus open-loop metrics and RFS. All four panels show the percentage of scenes with confirmed additional threats after binning trajectories by 3sADE, 5sADE, FDE, and RFS, respectively. 
  Non-trivial threat rates still persist in high-RFS regions, showing that these aggregate metrics do not fully capture safety-critical planning failures. See more cases in Appendix~\ref{sec:rfs-failures}.}
  \label{fig:open-loop-metrics-vs-threats}
  % \vspace{3pt}
\end{figure}

\begin{figure}[t]
  \centering
  \includegraphics[width=\linewidth]{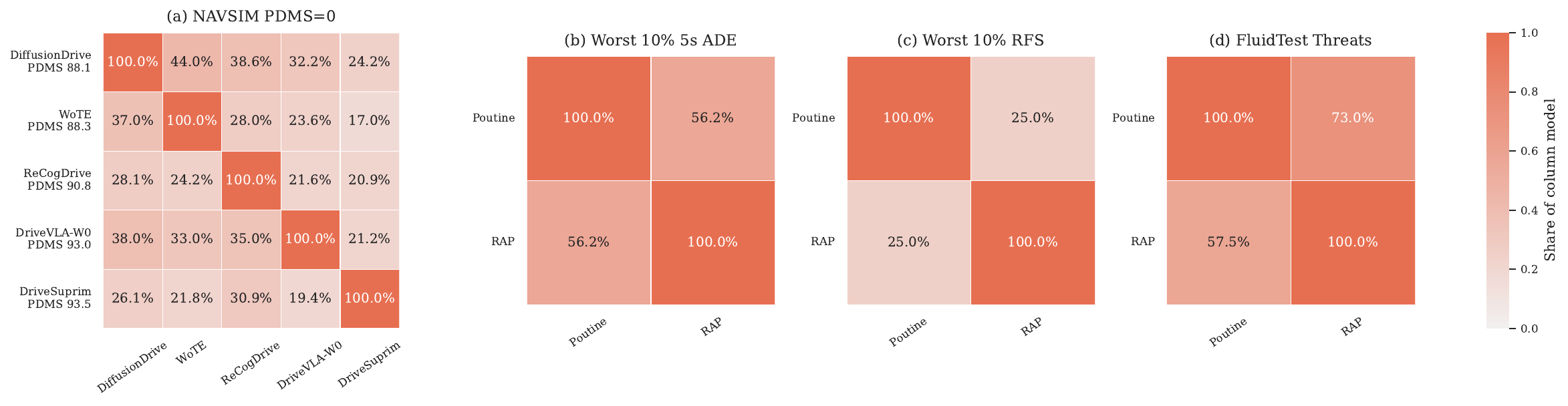}
  % \vspace{-10pt}
  \caption{Overlap analysis for long-tail planning evaluation. The heatmaps compare scenario sets selected by different metrics and planners, showing that existing metrics identify planner-dependent hard cases rather than stable long-tail failures.}
  \label{fig:overlap-heatmaps}
  % \vspace{1pt}
\end{figure}

\subsection{Human Annotation Protocol and WebUI}
\label{sec:human-annotation-webui}

We collect labels through pairwise additional-threat annotation. Instead of assigning an absolute score, annotators compare the expert trajectory $\tau_s^{*}$ and planner trajectory $\hat{\tau}_s$ under the same visual context and answer: \emph{Does the planner trajectory introduce any additional driving threat compared with the expert trajectory?} We find that the expert trajectory is essential for both stable human labels and reliable VLM-based explanations.

Annotators choose among three labels: \texttt{Y}, meaning the planner introduces at least one additional threat; \texttt{N}, meaning no additional threat despite possible geometric deviation; and \texttt{Not Sure}, used when future uncertainty, unexplained expert behavior, or noisy projection prevents a reliable binary decision. We treat \texttt{Not Sure} as abstention. Appendix~\ref{sec:webui-appendix} gives the full WebUI guidance, and Appendix~\ref{sec:human-label-cost} reports labeling-time statistics.

\subsection{Testing Protocol and FluidTest Safety Arena (Gold)}
\label{sec:testing-protocol}

FluidTest combines human labels with agentic verification and precision control. After binary labeling, the Prosecutor proposes candidate threat classes; the Detector verifies them using evidence-grounded decision graphs; and a reflection loop revises ambiguous cases before producing the final confirmed threat set. Abstained cases are sent to human reviewers for final review. We use Codex as the main agentic pipeline because it can invoke tools such as cropping and zooming when additional visual evidence is needed. The overall pipeline is shown in Fig.~\ref{fig:main-pipeline}; additional graph and Prosecutor details are provided in Appendix~\ref{sec:appendix-graphs} and Appendix~\ref{sec:appendix-prosecutor-qualitative}.

% The three-agent loop also serves as a reliability-checking system. Human labels provide the reference signal, while disagreement patterns expose weaknesses in different stages. High Judge abstention or rejection rates may indicate unstable human labels; repeated Judge reflections suggest insufficient Detector evidence; and systematic disagreement between Judge outputs and gold human labels signals failures in either the reasoning pipeline or annotation protocol. This redundancy improves auditability and robustness compared with relying on a single evaluator.

The three-agent loop follows a neuro-symbolic verification design. The Prosecutor uses VLM semantics to propose candidate threats, the Detector verifies them through structured decision graphs, and the Judge performs reflection and consistency checks. This mirrors prior LLM-symbolic systems that use LLMs for flexible semantic translation while delegating reasoning, execution, or validation to structured symbolic modules~\citep{pan2023logic,gao2023pal,liu2023llmp}. Tool use and reflection further improve auditability: Codex can gather additional visual evidence when needed, and the Judge can request more evidence or reject unsupported conclusions~\citep{yao2023react,shinn2023reflexion}.

FluidTest can be deployed locally for indicative evaluation. Controlled cross-planner comparison, however, requires a shared testing protocol. We therefore release a testing server where users submit planning results against a dynamic closed test set. For each submission, several human annotators will label additional threats, after which the full FluidTest pipeline will classify, explain, and score the planner. Results will be reported on the FluidTest Safety Arena (Gold) leaderboard.

\begin{table}[t]
\centering
\caption{
  Final FluidTest scores on val151 after human review. Entries are no-violation rates; higher is better except for 5sADE. \NATRShort{} denotes the overall no-additional-threat rate.
  The low numbers highlight a large gap between AI planners and human driving. 
}
\label{tab:grouped_metric_scores_human_review}
\resizebox{\textwidth}{!}{
\begin{tabular}{lcccccccc}
\toprule
Method 
& Quality~$\uparrow$
& Nav.~$\uparrow$
& Colli.~$\uparrow$
& Long.~$\uparrow$
& Prior.~$\uparrow$
& Lateral~$\uparrow$
& ADE~$\downarrow$
& NATR~$\uparrow$ \\
\midrule
Poutine~\citep{rowe2025poutine}
& 0.67 & 0.99 & 0.83 & 0.99 & 0.90 & 0.97 & \textbf{2.44} & 0.35 \\

VMA
& \textbf{0.72} & \textbf{1.00} & 0.84 & \textbf{1.00} & \textbf{0.93} & 0.97 & 2.46 & 0.47 \\

RAP~\citep{feng2025rap}
& 0.71 & 0.99 & \textbf{0.91} & 0.99 & 0.88 & \textbf{0.98} & 2.62 & \textbf{0.49} \\

% & Quality~$\uparrow$ 
% & Nav.~$\uparrow$ 
% & Colli.~$\uparrow$ 
% & Long.~$\uparrow$
% & Prior.~$\uparrow$ 
% & Lateral~$\uparrow$ 
% & 5sADE~$\downarrow$
% % & \colorbox{gray!20}{\strut 5sADE}
% & \textbf{\NATRShort{}~$\uparrow$} \\
% \midrule
% Poutine~\citep{rowe2025poutine}
% & 0.67
% & 0.84
% & 0.91
% & 0.97
% & \textbf{2.41} 
% % \colorbox{gray!20}{\strut 2.41}
% & 0.38 \\
% VMA
% & \textbf{0.75}
% & 0.82
% & \textbf{0.93}
% & 0.96
% & 2.46
% % \colorbox{gray!20}{\strut 2.46}
% & 0.47 \\
% RAP~\citep{feng2025rap}
% & 0.72
% & \textbf{0.90}
% & 0.88
% & \textbf{0.98}
% & 2.62
% % \colorbox{gray!20}{\strut 2.62}
% & \textbf{0.49} \\
\bottomrule
% \vspace{-10pt}
\end{tabular}
}
\end{table}

\begin{table}[t]
\centering
\caption{Threat-detection comparison for the Prosecutor. Without fine-tuning, large VLMs perform near random; supervised fine-tuning (SFT) results are evaluated on a 10\% test split.}
\label{tab:baseline_comparison}
\begin{tabular}{lcccccc}
% \hline
\toprule
\textbf{Model} & Dataset & \textbf{Acc} & \textbf{Precision} & \textbf{Recall} & \textbf{F1} & \textbf{F2} \\
\midrule
% \hline
Qwen3.5 9B & Poutine-Union
& 55\% & 67\% & 46\% & 54\% & 49\% \\
% ChatGPT 5.4-Mini medium
% & 49.45\% & 80.00\% & 18.11\% & 29.54\% & 21.42\% \\

ChatGPT 5.4-Mini xHigh & Poutine-Union
& 50\% & \textbf{100\%} & 8\% & 15\%  & 10\%\\

\textbf{Qwen3.5 4B SFT} & Poutine-Union
& \textbf{83\%} & 75\% & \textbf{100\%} & \textbf{85\%} & \textbf{93\%} \\

% \noalign{\vskip 2pt}
% \hline
% \noalign{\vskip 3pt}

\midrule

Qwen3.5 9B SFT & Poutine-1K
& 79\% & 85\% & 62\% & 71\% & 65\% \\

\bottomrule
% \vspace{1pt}
\end{tabular}
\end{table}

\section{Experiment Results and Insights}
\label{sec:exp}

We evaluate FluidTest through four questions. First, does the labeling protocol produce consistent human labels? Second, what do the final FluidTest scores reveal about current planners? Third, does a metric identify stable long-tail hard cases across planners? Fourth, can common open-loop and human-feedback metrics substitute for direct \NATRShort{} evaluation?

\subsection{Human Label Consistency and \NATRShort{} Learning}
\label{sec:human-label-consistency}

\paragraph{Human consistency requires task-specific prior knowledge.}
We ask five annotators, including three project researchers, to label the val151 subset with our WebUI for Poutine planning results. As shown in Fig.~\ref{fig:human_consistency_heatmaps}, the researchers produce more consistent labels than the non-researcher group. We find that non-researchers have difficulty reconstructing driving motion from projected trajectories, making it harder to anticipate future motion and identify potential threats. Within the researcher group, an overlap rate of \textbf{79.31\%} and a Fleiss' $\kappa$ value of \textbf{0.7145} indicate high agreement.

Importantly, this agreement is achieved under a deliberately challenging labeling protocol. High consistency is easy to obtain with narrow rules and hard boundaries. Instead, FluidTest asks annotators to conservatively identify all plausible additional threats relative to the expert trajectory, including potential future risks. This requires complex reasoning about all possible worst-case outcomes in defensive driving. 
% Labeling cost and interface details are provided in Appendices~\ref{sec:human-label-cost} and~\ref{sec:webui-appendix}.

\paragraph{Learning human safety preferences requires additional alignment.}
We use the Poutine union set for evaluation, defined as scenarios labeled \texttt{Y} by at least one researcher annotator. After excluding uncertain cases, both Qwen3.5-9B (no thinking) and ChatGPT-5.4-Mini xHigh perform close to random on the binary task of predicting additional threats. In contrast, a fine-tuned Qwen3.5 model performs substantially better, suggesting that human safety preferences require additional alignment, as shown in Table~\ref{tab:baseline_comparison} and more in Appendix~\ref{sec:appendix-prosecutor-qualitative}.

\subsection{Final FluidTest Benchmark Results}
\label{sec:final-fluidtest-results}
Table~\ref{tab:grouped_metric_scores_human_review} summarizes the final FluidTest results. The strongest 5sADE result does not imply the safest behavior: Poutine has the lowest 5sADE (2.44) but the lowest \NATRShort{} (0.35), meaning that 65\% of its trajectories introduce at least one additional threat. RAP has the worst 5sADE (2.62) but the best \NATRShort{} (0.49), while VMA ranks between them overall and has the best Quality and Priority scores. These results directly support our core claim: displacement quality and scalar human feedback do not reliably measure safety-relevant behavior in long-tail scenes.

\subsection{Reliable Hard-Case Metrics Should be Stable Across Planners}
\label{sec:hard-case-universality}

A useful long-tail benchmark should capture scenario-level difficulty rather than planner or metric related artifacts. Motivated by item-level evaluation theory and cross-model response analyses~\citep{martinez2019item,lalor2016evaluation,madhyastha2025taskaware,uzan2025charbench,vellamcheti2026cvt,kim2026euragovexam}, we evaluate \emph{failure-case universality}: whether hard cases selected by a metric are shared across planners.
Using random selectors as a baseline, the expected Jaccard overlap can be low. A metric that captures genuine scenario difficulty should identify substantially overlapping failure sets, even if the detailed failure modes differ.
% A random benchmark would produce near-zero worst-tail overlap across planners in the tail distribution; a metric that captures genuine scenario difficulty should identify substantially overlapping failure sets, even if detailed failure modes differ.

\paragraph{Closed-loop PDMS is weakly universal.}
We first evaluate five open-source state-of-the-art planners with diverse implementations on NAVSIM~\citep{liao2025diffusiondrive,li2025end,li2025recogdrive,li2025drivevlaw0,yao2026drivesuprim}. For each planner, we define failures as scenarios with $\mathrm{PDMS}=0$. As shown in Fig.~\ref{fig:overlap-heatmaps}, the overlap between planner failure sets is only \textbf{17\%} to \textbf{44\%}, suggesting that PDMS failures are strongly shaped by each planner's output distribution rather than by a shared set of universally difficult scenes.

\paragraph{RFS is less universal than open-loop displacement.}
Open-loop displacement provides an objective baseline for testing whether different planners share the same long-tail hard cases. A reliable long-tail metric should not produce lower cross-planner overlap than this baseline. On the val151 subset, the worst 10\% 5sADE hard tails of Poutine and RAP overlap by \textbf{56.2\%}. Repeating the analysis with Rater Feedback Score (RFS), using the worst 10\% of scenarios for each planner, yields only \textbf{25\%} overlap. This lower overlap indicates that RFS is a less stable hard-case selector, likely due to sparse labeling coverage and noise in scalar ratings.

\begin{figure}[t]
  \centering
  % \vspace{-10pt}
  \includegraphics[width=\linewidth]{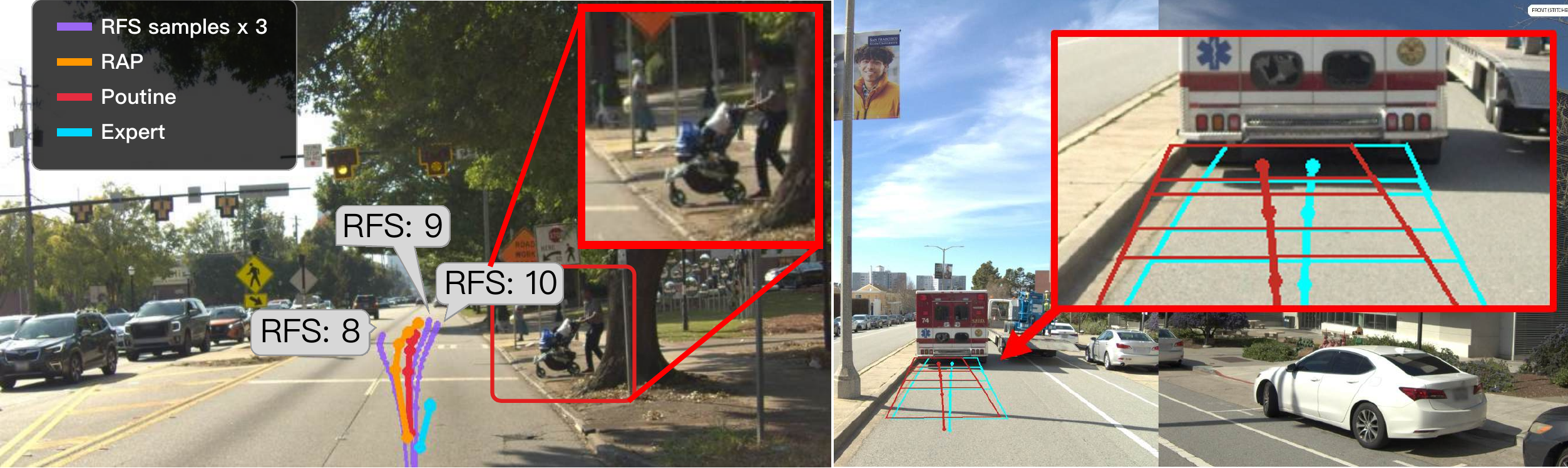}
  % \vspace{-10pt}
  \caption{
    RFS is not guaranteed to reflect critical safety hazards. Left: sparse labels do not cover the expert solution. Right: a small displacement in the wrong direction creates severe collision risk.}
  \label{fig:rfs_bad_case}
\end{figure}

\paragraph{\NATRShort{} is more reliable and informative.}

% Under \NATRShort{}, the directional overlap is \textbf{73.02\%} from Poutine-positive scenes to RAP-positive scenes and \textbf{57.50\%} in the reverse direction, with a Jaccard overlap of \textbf{47.42\%}.
Under \NATRShort{}, the \texttt{Y} sets overlap by \textbf{73\%}, substantially higher than the other metrics. This indicates that safety-relevant hard cases are largely shared across state-of-the-art planner families. Appendix~\ref{sec:overlap-analysis} reports threshold-sensitivity analyses showing that 5sADE and RFS need to cover about half of the dataset to reach comparable overlap, which no longer represents a hard tail.

Overall, low scores under existing metrics do not necessarily correspond to universal long-tail hard cases. Closed-loop PDMS and scalar RFS expose some failures, but their hard-case subsets remain heavily planner-dependent. FluidTest provides a more stable, interpretable, and planner-independent benchmark for long-tail planning safety.

\subsection{Relationship with Existing Metrics}
\label{sec:comparison-existing-metrics}

We examine whether existing open-loop displacement metrics and RFS can serve as proxies for direct additional-threat labels used to compute \NATRShort{}. For each pair, we partition the val151 scenarios into ten bins according to the metric value and report the percentage of human-labeled additional-threat cases in each bin. This analysis tests whether metric scores alone can recover the safety-relevant failures identified by FluidTest. The results show that no existing metric reliably predicts \NATRShort{} outcomes, especially when comparing across multiple planners.

\paragraph{Average displacement errors are limited safety indicators.}
Fig.~\ref{fig:open-loop-metrics-vs-threats} shows that 3sADE, 5sADE, and FDE are informative but limited. Displacement errors correlate with hazards within a single planner's output distribution, but the relationship weakens across planners. As shown in Fig.~\ref{fig:rfs_bad_case}, similar deviations can either create severe collision risk when they occur in the wrong direction or reflect a safe alternative mode. Thus, displacement errors are useful imitation-quality indicators but cannot serve as standalone safety metrics.

\paragraph{Longer-horizon displacement is more threat-sensitive.}
Among displacement metrics, 5sADE and FDE align better with additional threats than 3sADE. Many failures become visible only after the trajectory commits to a future interaction or route choice, such as entering a conflict zone, ending in an inappropriate lane, failing to prepare for a turn, or continuing toward an obstacle. Extreme displacements are therefore more reliably associated with additional threats.

\paragraph{Scalar human feedback does not reliably track additional threats.}
Although RFS uses human judgment, it performs worse than displacement for this purpose because of noise when comparing trajectories within and across planner distributions. Notably, the highest RFS bin still contains \textbf{40\%} additional threats, as shown in Fig.~\ref{fig:open-loop-metrics-vs-threats}. FluidTest instead directly evaluates whether a planner trajectory introduces additional risk relative to the expert trajectory.

\section{Conclusion}
\label{sec:conclusion}

Planning and testing in long-tail scenarios is difficult because such scenarios directly challenge the planners' generalization ability. Reliable evaluation should therefore remain grounded in human driving judgment while providing auditable structure. We introduced \textbf{FluidTest}, a human-aligned, safety-aware, verifiable, and explainable evaluation pipeline for long-tail autonomous driving planning. FluidTest reframes planning evaluation as additional-threat detection using expert behavior as a reference. It combines a pairwise WebUI for human annotation, a taxonomy of 32 safety-relevant threats, evidence-grounded decision graphs, and a Judge with reflection and revision. Across Poutine, RAP, and VMA, our results show that additional threats remain frequent and are often hidden by existing aggregate metrics. In contrast, \NATRShort{} provides a more stable and semantically meaningful view of long-tail planning safety.

\section{Limitations}
\label{sec:limitation}
% \paragraph{Limitations and Future Work.}

FluidTest has three main limitations. First, although we show that annotators can be trained to produce high-quality labels, the scalability of this procedure for labelers with diverse backgrounds remains to be tested. Second, the pipeline still depends on reliable trajectory projection, visual grounding, and sufficient scene context for human labeling. Third, although we can automatically expand the threat taxonomy from natural language guidelines, the current 32-threat taxonomy is not yet exhaustive.

% the current vanilla VLM Prosecutor may generalize unstably to unseen planner distributions; Appendix~\ref{sec:appendix-prosecutor-qualitative} reports more evidence and results for this issue. 

% Future work will scale FluidTest to more diverse datasets with richer temporal context, expand the taxonomy, and improve Prosecutor robustness through active learning and additional labeled data.

%%%%%%%%%%%%%%%%%%%%%%%%%%%%%%%%%%%%%%%%%%%%%%%%%%%%%%%%%%%%

\clearpage
% The acknowledgments are automatically included only in the final and preprint versions of the paper.
% \acknowledgments{If a paper is accepted, the final camera-ready version will (and probably should) include acknowledgments. All acknowledgments go at the end of the paper, including thanks to reviewers who gave useful comments, to colleagues who contributed to the ideas, and to funding agencies and corporate sponsors that provided financial support.}

%===============================================================================

\bibliography{example}  % .bib

\clearpage

%===============================================================================

\appendix

\section{WebUI for Annotators and Labeling Guidance}
\label{sec:webui-appendix}

\begin{figure}[t]
  \centering
  \includegraphics[width=1\textwidth]{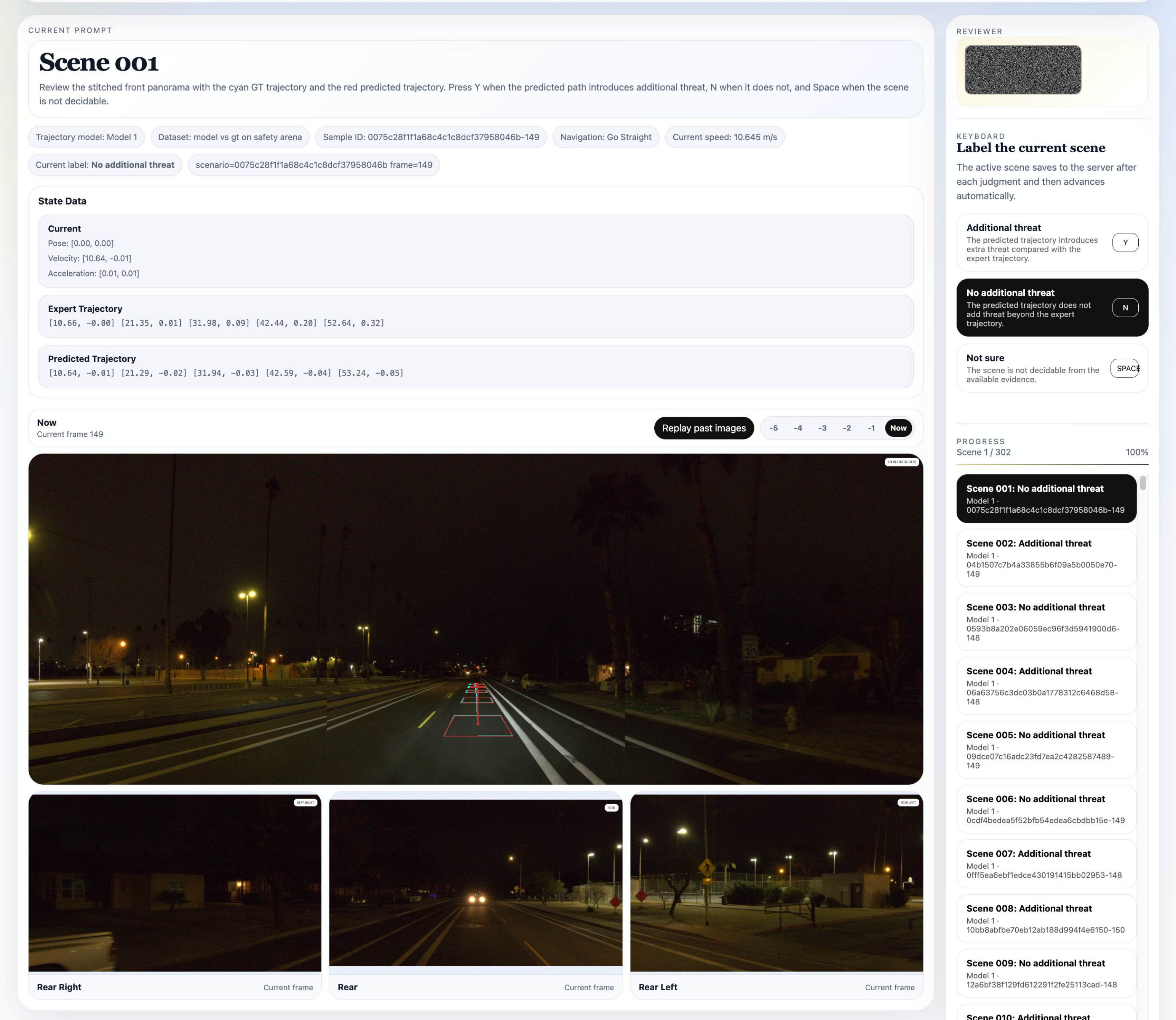}
  \caption{Screenshot of our WebUI for pairwise additional-threat annotation. The interface helps annotators compare the semantic consequences of two trajectories directly in image space.}
  \label{fig:webgui}
\end{figure}

We design a WebUI for annotators to label planning outputs and find that interface design is crucial for stable labels. Each annotator registers their name and driving experience, reads the labeling guidance, and then labels cases using the interface in Fig.~\ref{fig:webgui}. The interface includes scene information, trajectory-state data, a playback controller, three stitched front-camera images, and three rear-camera images. Playback covers the previous 1~s for all cameras. Lazy image loading and a magnifier improve labeling efficiency and accuracy.

The full guidance text is summarized below.

% \begin{description}[leftmargin=*, style=nextline]
\begin{description}[leftmargin=1em, labelindent=0pt, style=nextline]
  \item[Additional threat (\texttt{Y})]
  The predicted trajectory introduces an additional threat compared with the expert trajectory.

  \item[No additional threat (\texttt{N})]
  The predicted trajectory does not add any threat beyond the expert trajectory.

  \item[Not sure (\texttt{Space})]
  The scene cannot be judged from the available evidence.
\end{description}

\paragraph{How to label potential hazards.}
Annotators decide whether the prediction introduces anything meaningfully worse than the expert trajectory. Use \texttt{Y} only when the prediction creates an additional unnecessary maneuver or an additional potential hazard, including possible future risk. Use \texttt{N} when the prediction does not add anything worse than the expert trajectory. Use \texttt{Space}/Unsure when the expert trajectory has potential hazards but the available evidence is insufficient to understand why or compare fairly. The core reminder is to label added risk relative to the expert trajectory, not whether the scene itself is perfect.

\section{Human Label Cost}
\label{sec:human-label-cost}

\begin{figure}[t]
  \centering
  \includegraphics[width=0.85\textwidth]{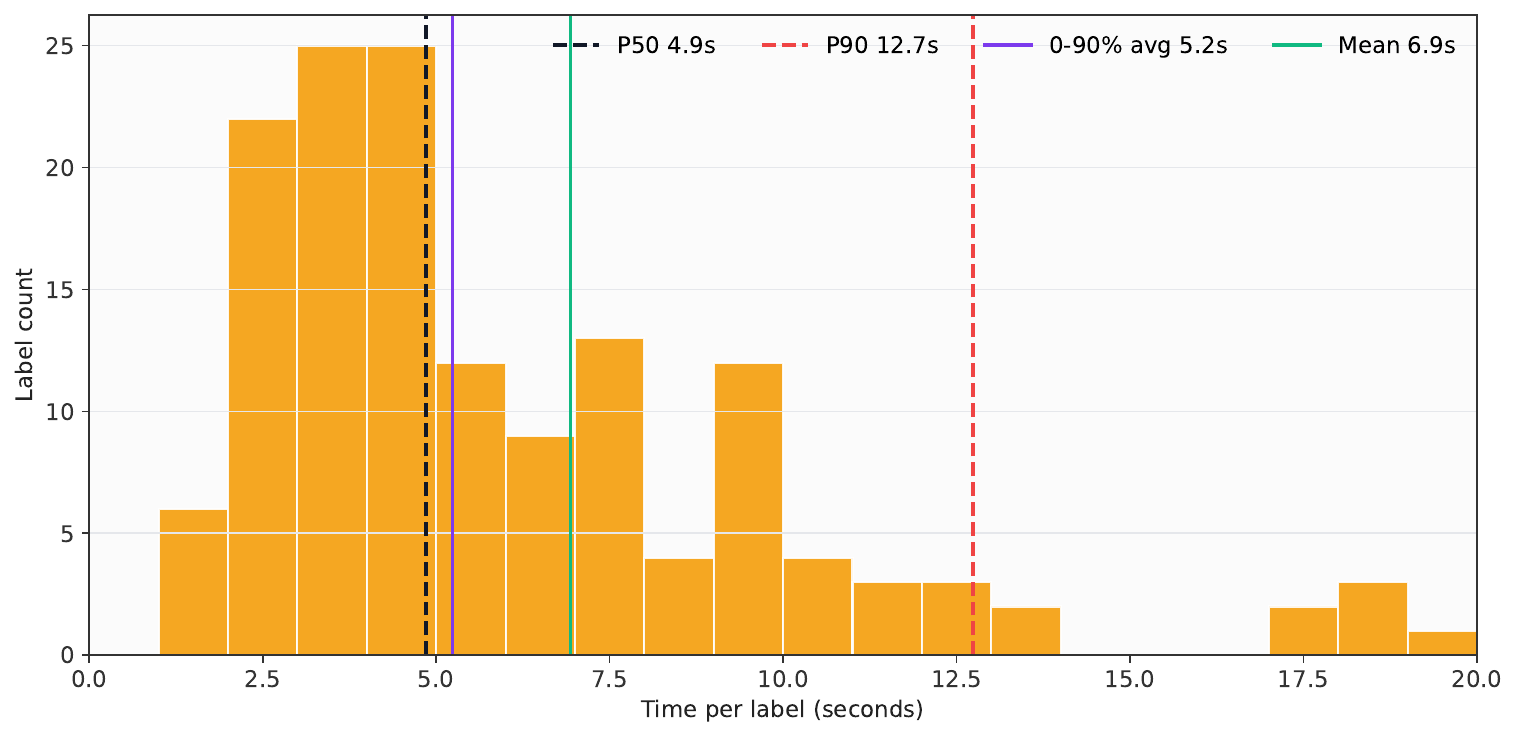}
  \caption{Detailed labeling-cost results measured by labeling time. The mean labeling time is 6.9~s and the median is 4.9~s, indicating that the interface supports efficient labeling for most long-tail scenarios. The labels were collected for VMA planned trajectories.}
  \label{fig:label-time-distribution}
\end{figure}

\paragraph{Labeling cost analysis.}
Previous experiments show that our \NATRShort{} metric is more reliable for human labelers than the previous RFS labels on the val151 subset.
In this section, we show that our labeling protocol requires minimal labeling effort.
To measure labeling cost, we log the approximate labeling time on the val151 subset of WOD-E2E. The average per-sample labeling time is about 5.292~s, meaning that labeling the full val151 subset requires less than 20 minutes. More details are shown in Fig.~\ref{fig:label-time-distribution}.

\paragraph{Training annotators for additional-threat labels.}
Additional-threat annotation requires task-specific calibration. Among the three researcher annotators, only Labeler B had direct knowledge of the labeling logic, the complete threat taxonomy and its definitions, and the Prosecutor classifier training pipeline.
The other two researcher annotators initially produced noisier labels, closer to those of the non-researcher annotators, especially for ambiguous cases requiring strict worst-case safety judgments.

We found that this gap was substantially reduced after expanding the labeling guidance with more representative examples. These examples clarified how to compare the planner trajectory against the expert trajectory, when to label plausible future risks as \texttt{Y}, and when to abstain because the expert behavior cannot be reliably explained. This suggests that consistent additional-threat labeling is trainable: annotators do not need to know the implementation details of FluidTest, but they do need calibrated examples that define the intended safety preference.

Due to time constraints, we did not perform the same training process for the other two non-researcher annotators. Therefore, the lower consistency observed in the non-researcher group should be interpreted as a limitation of our current annotation study rather than evidence that non-researchers cannot produce reliable labels. In the official FluidTest Safety Arena (Gold), we will include a standardized annotator-training stage with example-based guidance, calibration rounds, and consistency checks before collecting final benchmark labels.

\begin{figure}[t]
  \centering
  \includegraphics[width=1\textwidth]{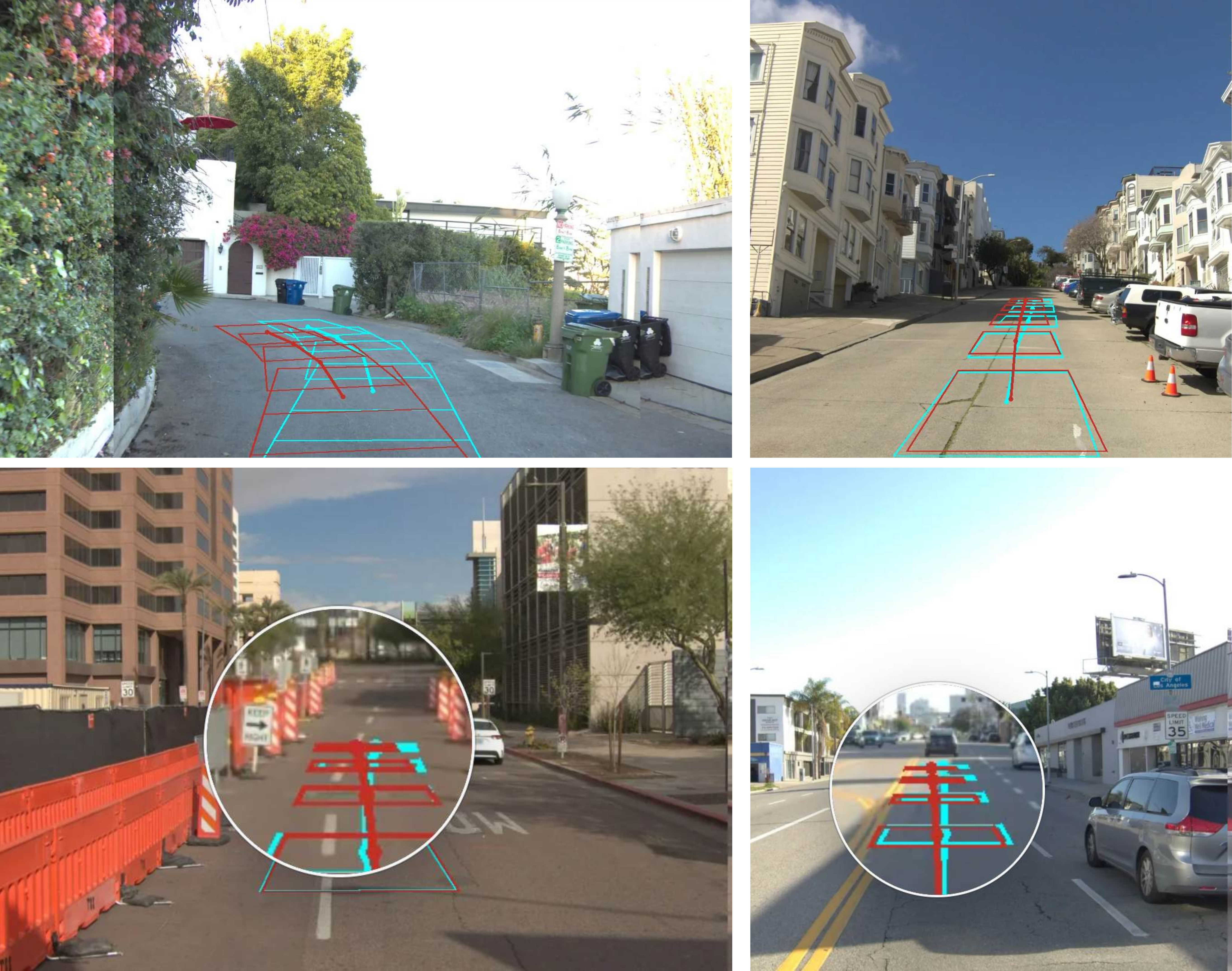}
  \caption{
    Representative annotator disagreement cases.
    In the top two examples, some reviewers interpreted the scene as a one-way road or a wide road without visible opposing traffic. Under that interpretation, they judged the planner's leftward deviation as not introducing an additional threat, even though the expert trajectory stays more clearly to one side of the road.
    In the bottom two examples, some reviewers judged the deviation as too small to create a threat or missed the deviation without carefully using the magnifier. The bottom-left case shows a leftward deviation toward traffic cones, while the bottom-right case shows a leftward deviation onto the double-yellow lines. In both cases, the planner introduces an additional threat that is not present in the expert trajectory.}
  \label{fig:disagree}
\end{figure}

\paragraph{Annotator disagreement cases.}
Human drivers vary in driving style and safety preference, ranging from conservative to aggressive. Annotators may also interpret the same scene differently, especially when road topology, lane direction, or obstacle clearance is ambiguous. In addition, labeling errors can occur when critical visual details are missed. Fig.~\ref{fig:disagree} shows four representative disagreement cases. These examples show that human annotation is not perfect even with our WebUI and labeling protocol. Therefore, reliable \NATRShort{} evaluation requires multiple trained annotators so that individual preferences, scene-interpretation differences, and occasional missed evidence do not dominate the final result.

\section{RFS Failure Cases}
\label{sec:rfs-failures}
\begin{figure}[t]
  \centering
  \includegraphics[width=1\textwidth]{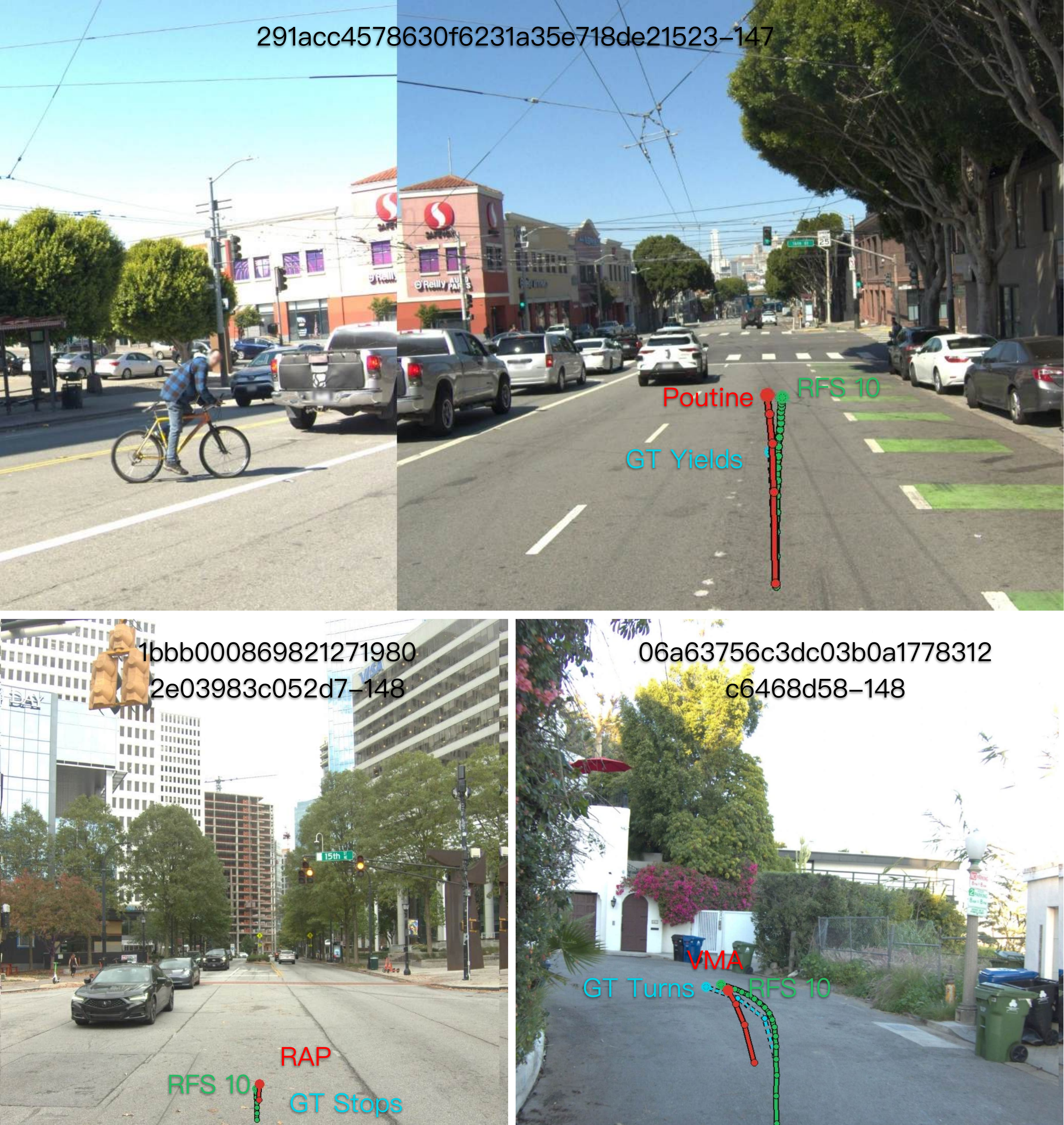}
  \caption{
    We show three poor planning results from the val151 subset that receive perfect RFS scores of 10 for different reasons across three planners. The cyan trajectory is the expert trajectory, 
    the green trajectory is the RFS10 reference trajectory, i.e., the maximum-score reference trajectory used by the RFS procedure for that scene, and the red trajectory is the planner output that receives an RFS of 10.}
  \label{fig:appendix-rfs-badcase}
\end{figure}

In this section, we provide additional case studies showing why RFS fails to fully capture long-tail driving threats.
WOD-E2E does not use the expert trajectory when computing RFS, aiming to cover multiple plausible future behaviors.
However, we argue that, in many cases, trajectories in the val151 subset with an RFS of 10 do not reflect this design goal.
We use ``RFS10 reference trajectory'' to denote the reference future trajectory used by the WOD-E2E RFS procedure that receives the maximum score of 10 for that scene.
In Fig.~\ref{fig:appendix-rfs-badcase}, we show three cases from three strong planners: Poutine, RAP, and VMA. The RFS values computed for these scenes are all 10, the maximum possible score.
In the upper case, however, both the RFS10 reference trajectory and the planner trajectory choose not to yield to the cyclist on the left, whereas the expert trajectory clearly yields; these two trajectories therefore introduce additional collision threats.
In the lower-left case, the expert trajectory remains stationary throughout the future 5-second planning horizon, while both the RFS10 reference trajectory and the RAP planner slowly move into the intersection, creating potential collision hazards with crossing vehicles.
In the lower-right case, the expert trajectory and the RFS10 trajectory turn left on the right side of the road. However, because the WOD-E2E RFS computation depends heavily on displacement and can fail on high-curvature trajectories, the poor VMA trajectory, which heads toward the road edge, still receives a perfect RFS of 10 despite clear additional threats.
These results further explain why the highest RFS bin in Fig.~\ref{fig:open-loop-metrics-vs-threats} still contains a substantial fraction of additional threats.

\section{Worst-Case Overlap Depends on the Tail Threshold}
\label{sec:overlap-analysis}

Section~\ref{sec:hard-case-universality} compares failure-case universality across different metrics using hard-case overlap rates. However, for continuous metrics such as 5sADE and RFS, the overlap rate depends strongly on how the long-tail threshold is defined. In the main text, we use the worst 10\% of scenarios to represent hard cases. Increasing the threshold naturally increases overlap because larger portions of the dataset are included.

\begin{figure}[t]
  \centering
  \includegraphics[width=1\textwidth]{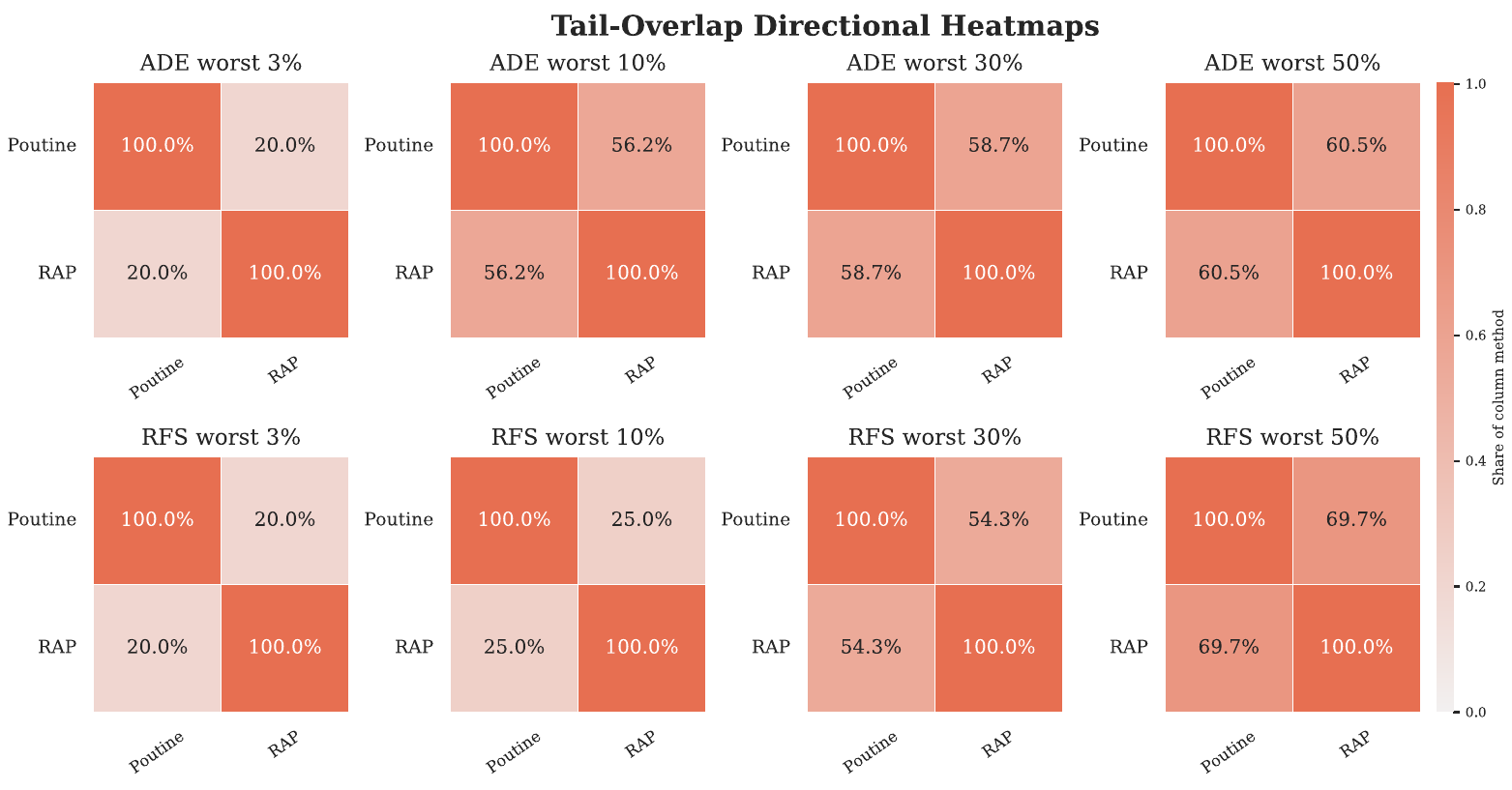}
  % \vspace{-4pt}
  \caption{
    Directional heatmap extending the previous hard-case overlap analysis. We use the Poutine model's predictions for this analysis to remain aligned with the previous analysis. The overlap rate of RFS increases after the threshold is raised above 30\%, indicating that RFS is unstable specifically on the 3--10\% tail and becomes more stable above 50\%.}
  \label{fig:more-overlap}
  % \vspace{4pt}
\end{figure}

To provide a fairer comparison, we report overlap rates under multiple thresholds in Fig.~\ref{fig:more-overlap} and Table~\ref{tab:overlap_analysis_table}. We observe that both 5sADE and RFS require covering roughly half of the dataset before reaching overlap rates comparable to \NATRShort{}. However, selecting 50\% of the dataset no longer meaningfully represents long-tail hard cases and provides limited diagnostic value for understanding why those scenarios are difficult.

\begin{figure}[t]
  \centering
  \includegraphics[width=1\textwidth]{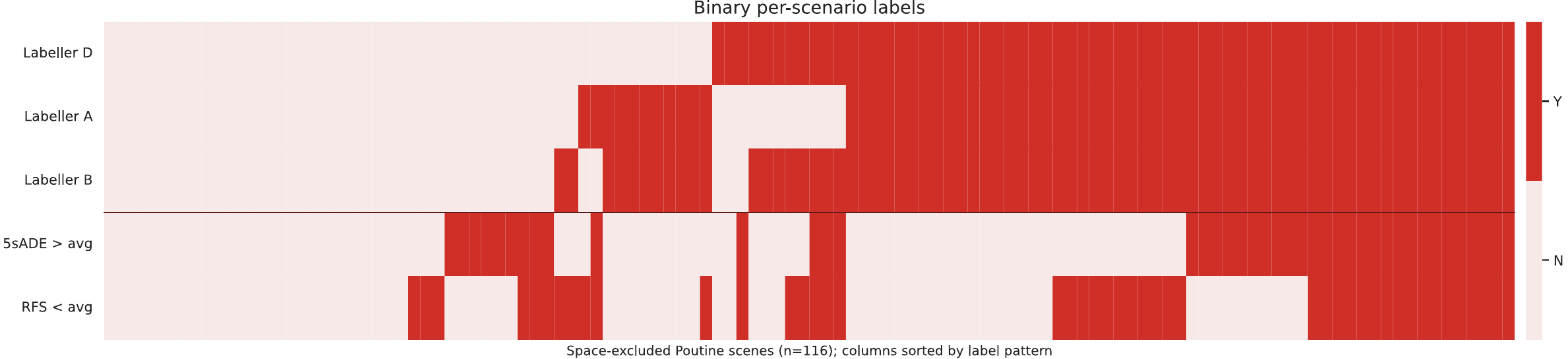}
  \caption{
    This per-scene heatmap shows that poor 5sADE and RFS scenarios do not necessarily produce more threats as measured by human labelers under \NATRShort{}. This result further emphasizes the value of human labeling with the proposed \NATRShort{} metric.}
  \label{fig:more-heatmap-on-y}
  % \vspace{4pt}
\end{figure}

We further ask whether 5sADE or RFS can identify scenarios where planners fail under \NATRShort{}. To study this question, we compare the worst 50\% of scenarios selected by each metric against human-labeled additional-threat cases. As shown in Fig.~\ref{fig:more-heatmap-on-y}, both 5sADE and RFS exhibit limited overlap with the human labeling results. 
This result further indicates that low 5sADE or RFS scores do not reliably correspond to human-identified safety failures under \NATRShort{}.

\begin{figure}[t]
  \centering
  \includegraphics[width=1\textwidth]{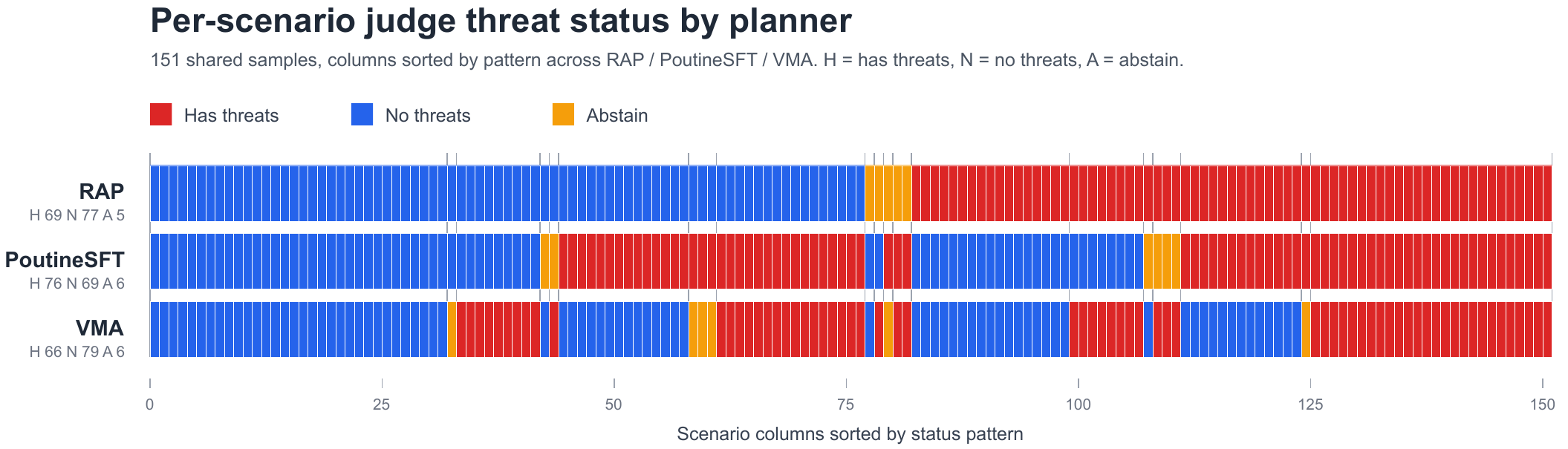}
  \caption{
    Per-scenario Judge results across all three planners on the val151 subset. The Judge abstentions indicate that the three-agent pipeline can resolve most cases and produce consistent explanations through the predefined decision graph.}
  \label{fig:appendix-judge-cases}
\end{figure}

\section{Prosecutor Threat Classifier}
\label{sec:appendix-prosecutor-qualitative}

Table~\ref{tab:baseline_comparison} shows that the threat-difference model reaches 79--83\% accuracy on the original evaluation sets. This is already strong because the target is not a hard-bounded geometric predicate, such as checking whether two trajectories overlap. Instead, the model must predict complex human safety preferences end-to-end.

The remaining errors mainly arise from preference ambiguity. Some cases are inherently ambiguous even for humans. After filtering out samples on which human annotators disagree, the post-filter evaluation becomes much more deterministic: the 2B variant reaches an accuracy of \textbf{0.929}, while the 4B variant reaches a perfect accuracy of \textbf{1.000} on the held-out evaluation set, as shown in Table~\ref{tab:threat-diff-predict-filter}. This suggests that much of the residual error in the unfiltered setting reflects ambiguity in the human preference labels rather than a failure to learn the main threat-difference rule.

\begin{table}[t]
\centering
\caption{Additional results for the Prosecutor additional-threat classifier. After filtering ambiguous cases in which the three human annotators disagree, the fine-tuned model achieves near-perfect prediction performance.}
\begin{tabular}{lllccc}
\toprule
Model & Acc. & Precision & Recall & F1 & F2 \\
\midrule
Qwen3.5 2B SFT & 0.7200 & 0.6429 & 0.8182 & 0.7200 & 0.7759 \\
Qwen3.5 4B SFT & 0.7200 & 0.6250 & 0.9091 & 0.7407 & 0.8333 \\
\midrule
Qwen3.5 2B SFT (filtered) & 0.9286 & 1.0000 & 0.8571 & 0.9231 & 0.8824 \\
Qwen3.5 4B SFT (filtered) & 1.0000 & 1.0000 & 1.0000 & 1.0000 & 1.0000 \\
\bottomrule
\end{tabular}
\label{tab:threat-diff-predict-filter}
\end{table}

\section{Judge Result Details}
\label{sec:appendix-judge-results}
We provide more detailed results from the Judge agent outputs. As shown in Fig.~\ref{fig:appendix-judge-cases}, the full pipeline can produce reliable and consistent evidence and decision-graph results, enabling the Judge to rule on most cases. Furthermore, the ratio of abstained cases remains largely consistent across planners, indicating that abstentions do not change the comparison results across different planners.

\section{Threat Definitions and Boundaries}
\label{sec:appendix-threats-def}

We provide the full definitions of all 32 threats in Table~\ref{tab:threat-definitions}. These definitions are used consistently by human annotators, result reviewers, and the agentic Codex server. See the accompanying documents and code for further details.

\begin{table}[t]
\centering
\caption{Scene overlap analysis between Poutine and RAP under different selection criteria.}
\label{tab:overlap_analysis_table}
\resizebox{\textwidth}{!}{%
\begin{tabular}{lcccc}
\toprule
% \textbf{Graph} & \textbf{Union scenes} & \textbf{Shared scenes} & \textbf{Poutine -> RAP} & \textbf{RAP <- Poutine} \\
\textbf{Criterion} & \textbf{Union scenes} & \textbf{Shared scenes} & \textbf{Shared / |RAP set|} & \textbf{Shared / |Poutine set|} \\
\midrule
5sADE worst 3\%  & 9  & 1  & 20.00\% & 20.00\% \\
5sADE worst 10\% & 23 & 9  & 56.25\% & 56.25\% \\
5sADE worst 30\% & 65 & 27 & 58.70\% & 58.70\% \\
5sADE worst 50\% & 106 & 46 & 60.53\% & 60.53\% \\
% 5sADE $< 1$      & 68 & 22 & 52.38\% & 45.83\% \\
RFS worst 3\%    & 9  & 1  & 20.00\% & 20.00\% \\
RFS worst 10\%   & 28 & 4  & 25.00\% & 25.00\% \\
RFS worst 30\%   & 67 & 25 & 54.35\% & 54.35\% \\
RFS worst 50\%   & 99 & 53 & 69.74\% & 69.74\% \\
\NATRShort{} (Ours)         & 97 & 46 & \textbf{73.02}\% & 57.50\% \\
\bottomrule
\end{tabular}%
}
\end{table}

\begin{table}[h]
\centering
\caption{Terminal states used by FluidTest threat graphs.}
\small
\begin{tabular}{lp{0.72\linewidth}}
\toprule
\textbf{State} & \textbf{Meaning} \\
\midrule
\texttt{confirmed} & Evidence sufficiently supports the threat. \\
\texttt{unproven} & Threat hypothesis is plausible but unsupported by sufficient evidence. \\
\texttt{lawful\_exception} & Behavior appears risky but is justified by traffic context or legal exception. \\
\texttt{routed} & Evidence better matches another sibling threat. \\
\texttt{not\_applicable} & Preconditions for this threat are absent. \\
\bottomrule
\end{tabular}
\label{tab:terminal-states}
\end{table}

%%%%%%%%%%%%%%%%%%%%%%%%%%%%%%%%%%%%%%%%%%%%%%%%%%%%%%%%%%%%

\begin{table*}[t]
  \centering
  \caption{Core reusable nodes shared across multiple threat graphs.}
  \label{tab:shared-nodes}
  \scriptsize
  \setlength{\tabcolsep}{4pt}
  \renewcommand{\arraystretch}{1.08}
  \begin{tabularx}{\textwidth}{p{0.24\textwidth} p{0.22\textwidth} X}
    \toprule
    \textbf{Node ID} & \textbf{Input} & \textbf{Function} \\
    \midrule

\texttt{NODE\_LANE\_OCCUPANCY}
& Trajectory + lane masks
& Determines whether the predicted trajectory occupies a valid lane region and measures lane overlap duration. \\

\texttt{NODE\_ROAD\_SIDE}
& Trajectory + roadway segmentation
& Determines whether the trajectory travels on the wrong roadway side or exits drivable space. \\

\texttt{NODE\_SIGNAL\_STATE}
& Traffic-light crop
& Detects signal state and movement permissions. \\

\texttt{NODE\_STOP\_CHECK}
& Ego velocity profile
& Determines whether the ego reaches a full stop before entering a controlled area. \\

\texttt{NODE\_PEDESTRIAN\_CONFLICT}
& Pedestrian detections + trajectory
& Detects whether the trajectory conflicts with pedestrian-priority zones. \\

\texttt{NODE\_VEHICLE\_CONFLICT}
& Vehicle tracks + trajectory
& Determines whether another vehicle possesses right-of-way conflict priority. \\

\texttt{NODE\_STATIC\_COLLISION}
& Static-object masks + trajectory
& Detects overlap or collision course with barriers, cones, curbs, poles, or other fixed objects. \\

\texttt{NODE\_STATIONARY\_VEHICLE}
& Parked-vehicle masks + trajectory
& Detects overlap or insufficient clearance to stationary vehicles. \\

\texttt{NODE\_FOLLOWING\_DISTANCE}
& Relative speed + spacing
& Estimates whether following distance is unreasonable for visible conditions. \\

\texttt{NODE\_LANE\_CHANGE}
& Temporal trajectory sequence
& Determines whether the ego commits to a lane change maneuver. \\

\texttt{NODE\_NAVIGATION\_MATCH}
& Navigation command + lane structure
& Checks whether the planner prepares correctly for a required turn or route branch. \\

\texttt{NODE\_SPEED\_REASONABLENESS}
& Ego speed + scene context
& Determines whether the speed is unsafe for visible environmental conditions. \\

\texttt{NODE\_WORKZONE}
& Cone/barrier segmentation
& Detects temporary work-zone narrowing and clearance risk. \\

\texttt{NODE\_TEMP\_CONTROL}
& Flagger/officer/cone cues
& Detects temporary traffic-control instructions and diversion paths. \\

    \bottomrule
  \end{tabularx}
\end{table*}

\begin{table*}[p]
\centering
\scriptsize
\setlength{\tabcolsep}{3pt}
\renewcommand{\arraystretch}{0.96}
\begin{tabularx}{\textwidth}{p{0.18\textwidth} p{0.29\textwidth} X}
\toprule
\textbf{Group} & \textbf{Threat label} & \textbf{Definition} \\
\midrule
Signal/stop control & Red light turning no full stop & Vehicle makes a turn on red without first coming to a full stop where the red-turn movement is allowed only after stopping and yielding. \\
Signal/stop control & Red light violation & Vehicle enters or proceeds through a steady red signal or makes a prohibited movement on red. \\
Signal/stop control & Stop sign no full stop & Vehicle fails to come to a full stop where a stop is required. \\
Signal/stop control & Yellow phase noncompliance & Vehicle enters on a steady yellow when a normal safe stop before the controlled entry was reasonably available. \\

Right-of-way & Failure to yield pedestrian & Vehicle fails to yield to a pedestrian in a marked or unmarked crosswalk or equivalent pedestrian priority zone. \\
Right-of-way & Failure to yield vehicle & Vehicle enters, merges, or turns across another vehicle's or bicyclist's right-of-way unsafely. \\

Collision & Collision with static objects & Predicted ego travel remains on a path that would strike, scrape, or is clearly driving into a plausible future collision course with, a static object; judge the ego swept footprint/corridor, not only the centerline overlay, and visible overlap within the current 5s path is not required when ego is aimed toward the object and GT is not. \\
Collision & Collision with stationary vehicle & Predicted ego travel remains on a path that would strike a parked or otherwise stationary vehicle. \\
Collision & Vehicle collision course & Predicted ego travel remains on a path that would strike another vehicle unless one of the vehicles changes course, outside narrower right-of-way, tailgating, or lane-change explanations. \\

Following distance & Following too closely & Vehicle follows another vehicle more closely than is reasonable and prudent for the visible speed and traffic context. \\

Lane use & Lane straddling & Vehicle drives for a sustained period between two lanes instead of committing to one lane. \\
Lane use & Unsafe lane change & Vehicle changes lanes without reasonable safety. \\
Lane use & Unsafe passing & Vehicle passes in a prohibited or unsafe manner. \\
Lane use & Wrong side of road & Vehicle travels on the oncoming or left side of the roadway outside a clearly lawful exception. \\
Lane use & Wrong way one way road & Vehicle travels opposite the designated direction on a clearly one-way roadway. \\

Roadway position & Off road driving & Vehicle leaves the ordinary drivable lane or roadway and drives on a sidewalk, curb area, shoulder, dirt, gore, median edge, parking apron, private-property drive/frontage area, landscaped area, or another non-roadway area without a clearly lawful or necessary explanation. \\

Special-lane misuse & Driving in bike lane & Vehicle uses a bicycle lane as a normal travel lane beyond narrow lawful exceptions. \\
Special-lane misuse & Driving in bus only lane & Vehicle uses a lane reserved for public transit buses outside allowed conditions. \\
Special-lane misuse & Improper shoulder pass or bypass & Vehicle uses the shoulder, emergency lane, or other off-main-traveled roadway edge as a pass, bypass, or ordinary travel path outside clear lawful exceptions. \\

Speed & Speeding posted limit & Vehicle clearly exceeds a posted speed limit when the limit and excess are visually well supported. \\
Speed & Unsafe speed for conditions & Vehicle is traveling faster than is reasonable for the visible scene conditions. \\

Reckless pattern & Speed contest or exhibition & Vehicle behavior suggests racing, competitive acceleration, or exhibition of speed. \\
Pattern & Aggressive weaving & Vehicle repeatedly changes lanes or swerves laterally in a rapid, conflict-seeking, or intimidation-like manner that materially raises interaction risk. \\

Route compliance & Late navigation lane preparation & Vehicle stays out of the lane position needed for a near-term instructed turn, exit, or branch until the last reasonable preparation chance is closing or already missed, making future route failure or a forced risky late lane change likely. \\
Route compliance & Not following navigation instruction & Vehicle materially deviates from a clear near-horizon navigation instruction by committing to the wrong turn, branch, or lane stack without a visible reason. \\

Temporary control & Not following temporary traffic instructions & Vehicle materially disregards a clear temporary traffic-control instruction from an officer, flagger, construction guide, cone-directed diversion, or similar active traffic-direction setup. \\

Policy quality & Blocking traffic without necessity & Vehicle stops or crawls in a way that unnecessarily obstructs a lane, crosswalk, or intersection without a legitimate traffic-control reason. \\
Policy quality & Low efficiency driving & Vehicle remains static or crawls at an obviously inefficient pace without a visible traffic-control, safety, conflict, or route-based reason. \\
Policy quality & Meaningless lane change & Vehicle changes lanes without a reasonable traffic-related purpose or weaves left-right-left without improving progress. \\
Policy quality & Meaningless lateral drift & Compared with GT, the predicted vehicle adds or retains material lateral deviation without a clear purpose, while not fully becoming a lane change, lane straddling, or roadway departure. \\
Policy quality & Noncommittal merge & Vehicle begins a merge or turn commitment and then hesitates or aborts in a way that reflects noncommittal go-no-go behavior and may disrupt surrounding traffic. \\
Policy quality & Stops without escape room & Vehicle stops too close to an obstacle or blocked area and leaves insufficient space for a normal forward go-around if the obstacle remains stationary. \\
\bottomrule
\end{tabularx}
\caption{Compact definitions of the semantic threat labels used by FluidTest.}
\label{tab:threat-definitions}
\end{table*}

%%%%%%%%%%%%%%%%%%%%%%%%%%%%%%%%%%%%%%%%%%%%%%%%%%%%%%%%%%%%

\section{Threat Decision Graphs and Reusable Nodes}
\label{sec:appendix-graphs}

\begin{figure*}[t]
\centering
\resizebox{\textwidth}{!}{%
\begin{tikzpicture}[
  font=\footnotesize,
  node distance=7mm and 13mm,
  >=Latex,
  decision/.style={
    draw,
    rounded corners=2pt,
    align=center,
    text width=38mm,
    inner sep=4pt,
    fill=blue!4
  },
  terminal/.style={
    draw,
    rounded corners=2pt,
    align=center,
    text width=30mm,
    inner sep=4pt,
    fill=gray!10
  },
  confirmed/.style={
    draw,
    rounded corners=2pt,
    align=center,
    text width=30mm,
    inner sep=4pt,
    fill=green!12
  },
  routed/.style={
    draw,
    rounded corners=2pt,
    align=center,
    text width=34mm,
    inner sep=4pt,
    fill=orange!12
  },
  edge/.style={->, line width=0.45pt},
  lab/.style={midway, fill=white, inner sep=1pt}
]

\node[decision] (n0) {N1. Completed lane change visible?};
\node[decision, below=of n0] (n1) {N2. Better explained as vehicle right-of-way taking?};
\node[decision, below=of n1] (n2) {N3. Better explained as aggressive weaving?};
\node[decision, below=of n2] (n3) {N4. Inadequate safety margin to adjacent traffic?};
\node[decision, below left=13mm and 12mm of n3] (n4) {N5. Forced reaction, near miss, or collision?};
\node[decision, below right=13mm and 12mm of n3] (n5) {N6. Legitimate merge, route, obstacle, or emergency reason?};
\node[decision, below=of n5] (n6) {N7. Better explained as noncommittal merge?};
\node[decision, below=of n6] (n7) {N8. Lane change lacks traffic or route purpose?};

\node[confirmed, below=14mm of n4] (tc) {Confirmed\\\texttt{unsafe\_lane\_change}};
\node[routed, right=28mm of n1] (tr1) {Routed\\\texttt{failure\_to\_yield\_vehicle}};
\node[routed, right=28mm of n2] (tr2) {Routed\\\texttt{aggressive\_weaving}};
\node[routed, right=28mm of n6] (tr3) {Routed\\\texttt{noncommittal\_merge}};
\node[routed, right=28mm of n7] (tr4) {Routed\\\texttt{meaningless\_lane\_change}};
\node[terminal, left=26mm of n0] (tna1) {Not applicable};
\node[terminal, left=26mm of n5] (tna2) {Not applicable};
\node[terminal, left=26mm of n2] (tu1) {Unproven};
\node[terminal, left=26mm of n7] (tu2) {Unproven};

\draw[edge] (n0) -- node[lab] {yes} (n1);
\draw[edge] (n0) -- node[lab] {no} (tna1);
\draw[edge] (n0.west) to[out=190,in=90] node[lab] {abstain} (tu1.north);

\draw[edge] (n1) -- node[lab] {no} (n2);
\draw[edge] (n1) -- node[lab] {yes} (tr1);
\draw[edge] (n1.west) to[out=180,in=20] node[lab] {abstain} (tu1);

\draw[edge] (n2) -- node[lab] {no} (n3);
\draw[edge] (n2) -- node[lab] {yes} (tr2);
\draw[edge] (n2.west) -- node[lab] {abstain} (tu1);

\draw[edge] (n3) -- node[lab] {yes} (n4);
\draw[edge] (n3) -- node[lab] {no} (n5);
\draw[edge] (n3.west) to[out=180,in=-20] node[lab] {abstain} (tu1);

\draw[edge] (n4) -- node[lab] {yes/no/abstain} (tc);

\draw[edge] (n5) -- node[lab] {yes} (tna2);
\draw[edge] (n5) -- node[lab] {no} (n6);
\draw[edge] (n5.west) to[out=180,in=120] node[lab] {abstain} (tu2);

\draw[edge] (n6) -- node[lab] {yes} (tr3);
\draw[edge] (n6) -- node[lab] {no/abstain} (n7);

\draw[edge] (n7) -- node[lab] {yes} (tr4);
\draw[edge] (n7) -- node[lab] {no} (tna2);
\draw[edge] (n7.west) -- node[lab] {abstain} (tu2);
\end{tikzpicture}%
}
\caption{Full authored decision graph for \texttt{unsafe\_lane\_change}. The graph first proves a completed lane change, routes cases better explained by vehicle right-of-way taking or aggressive weaving to narrower sibling threats, confirms unsafe lane change when inadequate margin is present, and suppresses cases explained by legitimate route, merge, obstacle, or emergency needs.}
\label{fig:unsafe-lane-change-decision-graph}
\end{figure*}
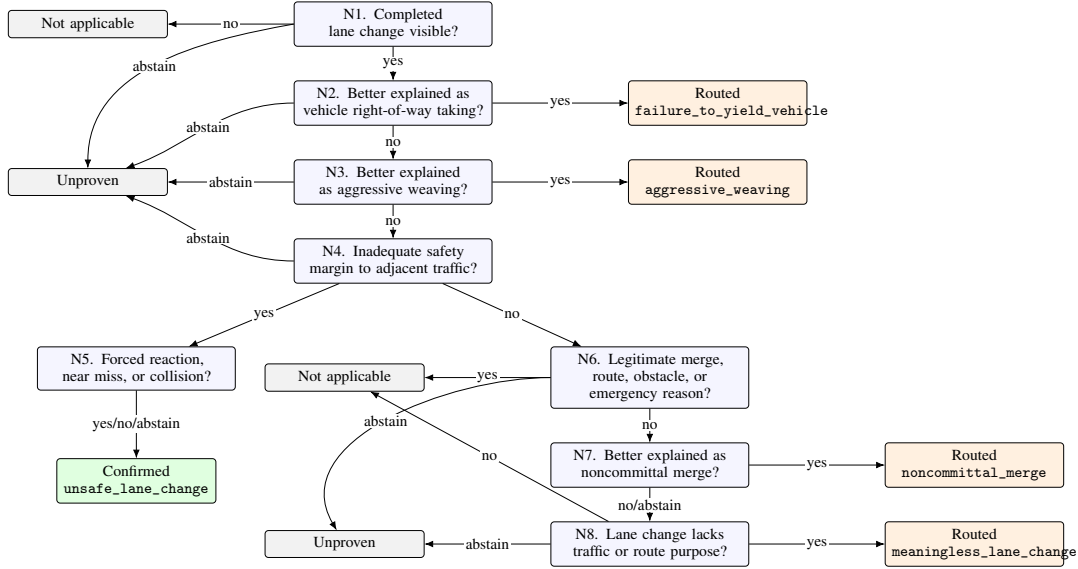

We refer readers to this \href{https://safety-arena-web.vercel.app/threat-graph}{online site} for detailed decision graphs for all threats, since the full graphs are too large to include in the paper. Each semantic threat is implemented as a structured graph composed of reusable evidence nodes, routing logic, lawful-exception checks, and terminal decision states. 
We include one representative graph in full and summarizes the terminal conditions for all authored
threat graphs used by the evaluator, as shown in Fig.~\ref{fig:unsafe-lane-change-decision-graph}.

\subsection{Graph Design Principles}

Each threat graph follows four principles:

\begin{enumerate}[leftmargin=*]
    \item \textbf{Evidence grounding.}
    Every accepted threat must be supported by explicit observable evidence from trajectory geometry, scene context, visual grounding, or temporal behavior.

    \item \textbf{Composable reusable nodes.}
    Shared checks such as lane occupancy, signal state, pedestrian conflict, collision overlap, and roadway-side validation are implemented once and reused across multiple threat graphs. Details are provided in Table~\ref{tab:shared-nodes}.

    \item \textbf{Auditable execution.}
    The graph stores intermediate node outputs, queried visual evidence, routing decisions, and lawful-exception reasoning.

    \item \textbf{Multi-terminal reasoning.}
    Graphs do not produce only binary outputs. Possible terminal states include \texttt{confirmed}, \texttt{unproven}, \texttt{lawful\_exception}, \texttt{routed}, and \texttt{not\_applicable}.
\end{enumerate}

\subsection{Terminal States}
As shown in Table~\ref{tab:terminal-states}, we define five terminal states for each threat graph. The \texttt{unproven}, \texttt{lawful\_exception}, and \texttt{not\_applicable} states all correspond to different forms of ``no threat'' for the currently evaluated threat. The \texttt{routed} state triggers an additional graph check for the routed threat. These final decisions are then provided to the Judge for the final determination.

\end{document}